\documentclass[11pt]{article}
\usepackage{eamt24}
\usepackage{times}
\usepackage{url}
\usepackage{latexsym}
\usepackage[small,bf]{caption} 
\title{Constructing and Expanding Low-Resource and Underrepresented Parallel Datasets for Indonesian Local Languages}

\author{Joanito Agili Lopo\\
  Universitas Gadjah Mada\\
  {joanitoagililopo@mail.ugm.ac.id}  \And
  Radius Tanone\\
  Universitas Kristen Satya Wacana\\
  {radius.tanone@uksw.edu}}

\date{}

\begin{document}
\maketitle
\begin{abstract}
  In Indonesia, local languages play an integral role in the culture. However, the available Indonesian language resources still fall into the category of limited data in the Natural Language Processing (NLP) field. This is become problematic when build NLP model for these languages. To address this gap, we introduce \textbf{Bhinneka Korpus}\footnote{https://github.com/joanitolopo/bhinneka-korpus}, a multilingual parallel corpus featuring five Indonesian local languages. Our goal is to enhance access and utilization of these resources, extending their reach within the country. We explained in a detail the dataset collection process and associated challenges. Additionally, we experimented with translation task using the IBM Model 1 due to data constraints. The result showed that the performance of each language already shows good indications for further development. Challenges such as lexical variation, smoothing effects, and cross-linguistic variability are discussed.  We intend to evaluate the corpus using advanced NLP techniques for low-resource languages, paving the way for multilingual translation models.
\end{abstract}

\section{Introduction}

Indonesia, the world's fourth most populous nation with 273 million people spread across 17,508 islands, is also the second most linguistically diverse country after Papua New Guinea, accounting 10\% of the world's total languages. This linguistic richness, with 718 identified local languages from 1991 to 2019, excluding dialects and sub-dialects, underscores the vital role of indigenous languages in conveying cultural values within the community \cite{Alamsyah2018/01}. 

Natural Language Processing (NLP) plays a crucial role in preserving this linguistic diversity and cultural heritage \cite{cahyawijaya-etal-2023-nusawrites}, yet building effective NLP models requires access to sufficient resources, a challenge for Indonesian categorized as a language with limited resources. Nevertheless, recent initiatives by NLP researchers are making strides in gathering Indonesian local language resources, marking progress in addressing resource scarcity and advancing NLP technology in Indonesia \cite{cahyawijaya-etal-2023-nusawrites,winata-etal-2023-nusax,Prabowo_Gabriel_Nazarudin_Ratumanan_Maslim_2024,koto-koto-2020-towards,7065828,cahyawijaya-etal-2021-indonlg,6709907}.

\begin{figure*}
 \centering
 \includegraphics[scale=0.5]{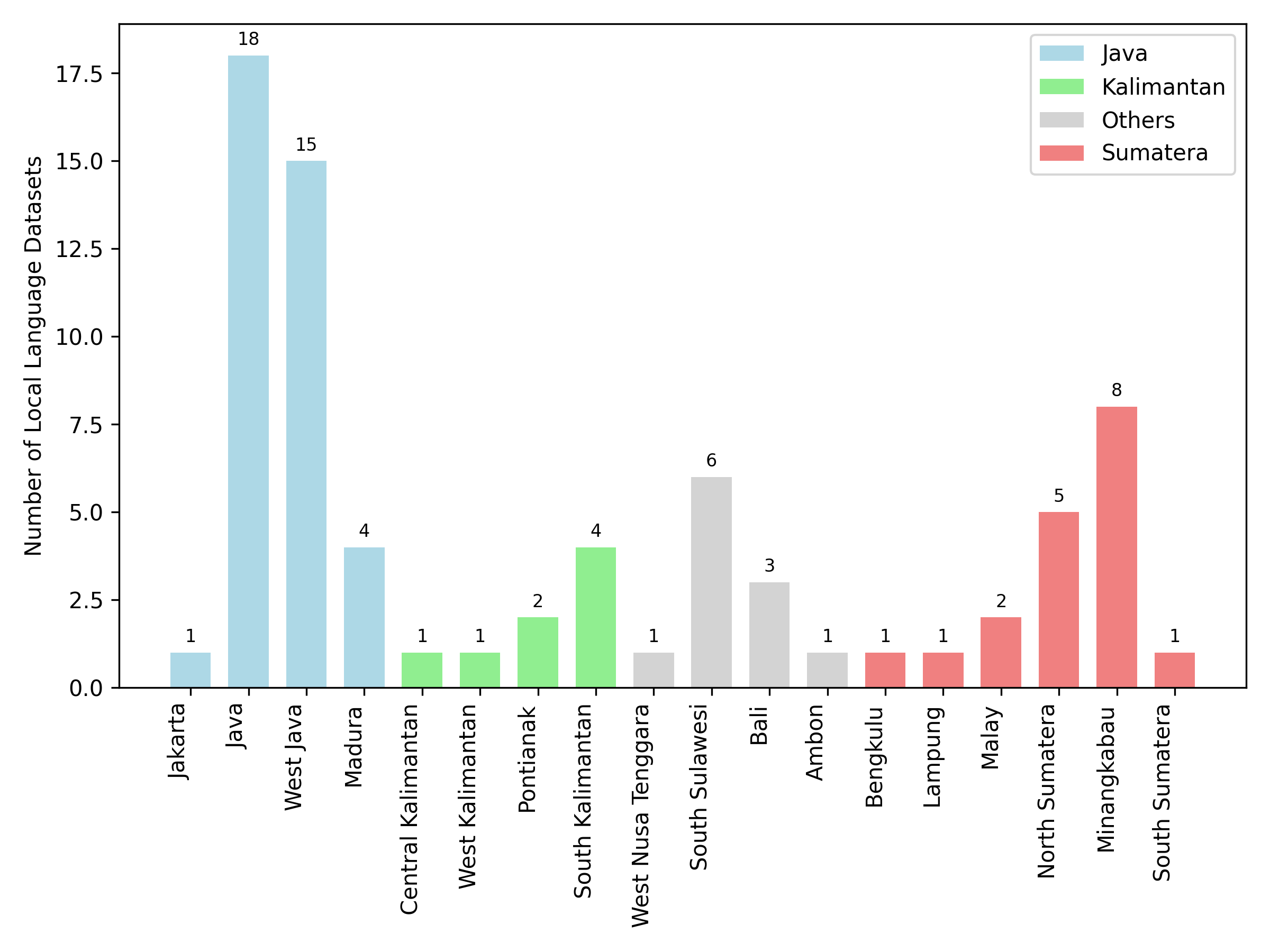}  
 \caption{Number of open datasets available for Indonesian local languages}
\end{figure*}

Unfortunately, NLP resources for Indonesian local languages are currently biased towards Western Indonesia, neglecting linguistic nuances and variations in other regions and dialects. Factors such as poorly recorded in written works, less officially taught, have a digital divide and isolation due to cultural barriers, and do not have guidelines that are applied in the country, hindering NLP research in these languages \cite{amien2022locationbased,aji-etal-2022-one,novitasari-etal-2020-cross}. Figure 1 illustrates the scarcity of open datasets for Indonesian local languages, with Java and Sumatra having the most resources, followed by Kalimantan and others with fewer\footnote{Our reference based on open dataset by IndoNLP Data Catalog \cite{cahyawijaya-etal-2023-nusacrowd}}. Despite Indonesia's 700+ languages, it is just about 2.5\% compared with the available resources. Eastern and Central Indonesia, despite fewer speakers, boast more diverse languages, with Papua, Maluku, East Nusa Tenggara, and Sulawesi alone comprising over 70 languages, around 89.41\% of the total identified in Indonesia.

To overcome the gap above, we introduce Bhinneka Korpus, a multilingual parallel corpus for five Indonesian local languages. Our primary focus lies on central and eastern Indonesian languages and present the first bilingual lexicon for an under-documented local language from West Kalimantan, spoken by fewer than 5,000 people. We also describe an approach and challenge to collecting low-resource local language dataset. We aim to develop more inclusive and culturally relevant NLP solutions. 

\begin{table*}[t]
\centering
\begin{tabular}{llll}
\hline
\textbf{Language} & \textbf{Speakers} & \textbf{Annotators’ Dialect} & \textbf{Example} \\
\hline
Ambonesse (abs) & 200,000 & Ambon Malay & Beta nanti pi deng se \\
Beaye (day) & $\sim$5,000 & Beaye & Ken yak ampus ngan ko \\
Kupang Malay (mkn) & 200,000 & Basa Kupang & Nanti beta pi deng lu \\
Makassarese (mak) & 1,870,000 & Maros-Pangkep & Naklampaya segang kau \\
Uab Meto (aoz) & 700,000 & Amanuban-Amanatun, Mollo-Miomafo & Au he nao ok ko \\
\hline
\end{tabular}
\caption{Language Information}
\end{table*}

\section{Related Work}

Indonesia, being recognized as one of the world's most linguistically diverse countries, has seen extensive language research. Numerous scientists have taken on the task of creating local language datasets, such as NusaX \cite{winata-etal-2023-nusax}, a multilingual sentiment dataset for 10 Indonesian languages. Similarly, Aji et al \shortcite{aji-etal-2022-one} provided a condition of NLP research in Indonesia, focusing on its diverse linguistic landscape of over 700+ languages. This study underscores challenges in Indonesian NLP research and the need for further investigation of underrepresented languages.

Despite Indonesia's vast population, its linguistic data remains limited, falling into the category of resource-poor data, and faces inherent challenges in the NLP context. Researchers like Cahyawijaya et al \shortcite{cahyawijaya-etal-2023-nusawrites} initiated investigations, resulting in the development of NusaWrites, focusing on Indonesia's local languages. Their study found that datasets created by native speakers through paragraph writing exhibited higher quality in terms of lexical diversity and cultural content. Additionally, the use of location-based Twitter filtering techniques, as demonstrated by Amien et al~\shortcite{amien2022locationbased}, could be used for creating, collecting, and categorizing local Indonesian datasets for NLP and addressing resource limitations in datasets.

However, the collected process of local languages in Indonesia reveals shared dialectal similarities like Malaysia, which is leading to NLP challenges such as code-mixing. Thus, Adilazuarda et al \shortcite{adilazuarda-etal-2022-indorobusta} introduced the IndoRobusta framework to assess and enhance code-mixing resilience. Their study delves into code mixing in Indonesian, particularly in four embedded languages: English, Sundanese, Javanese, and Malay.

The insufficient of local languages in Indonesia poses a challenge for NLP research. To address this, it is quite reasonable to apply machine learning and deep learning techniques to address language-specific challenges in Indonesia. In Koto and Koto \shortcite{koto-koto-2020-towards} research, they conduct the Minangkabau language using traditional machine learning methods and sequence-to-sequence models like LSTM and Transformer respectively. Initial experiments showed a notable decrease in Minangkabau text classification accuracy when using a model trained on Indonesian data. In their experiment, a basic word-to-word translation approach using a bilingual dictionary outperformed LSTM and Transformer models in terms of BLEU score.

This study presents a novel contribution to the effort of curating low resource machine translation dataset. We focused on five underrepresented Indonesia local languages with main goals is to enhance the existing Indonesian local language resources specifically on central and eastern Indonesian languages.

\section{Languages Overview}
We collected Indonesian-Ambon Malay, Kupang Malay, Beaye, Makassarese, and Uab Meto parallel corpus, and releases the first bilingual lexicon for Beaye. Table 1 provides information of languages dialects, including their language codes, approximate number of speakers, the specific dialect spoken by annotators, and a sample sentence showcasing each language's characteristics. For instance, Ambonesse Malay, represented by the code 'abs', is spoken by around 200,000 first language speakers, with annotators using the Ambon Malay dialect. Meanwhile, Figure 2 shows each of the language taxonomy. Following section will be explain in a brief about these languages. 

\subsection{\textbf{Ambonese Malay}} 
Ambonese Malay (abs) is a language spoken in the central and southern Moluccas Islands in Eastern Indonesia and today serves as a lingua franca for market interaction and media communication. It is written in Latin script and has been known since the 14th century \cite{Suharyanto2015} and developed directly from the Old Malay language \cite{collins2005bahasa}. However, because of the influence from other Malay and external sources such as Dutch and Portuguese colonial, made the language differ from the Old Malay and absorbed numerous lend words \cite{Muskita2022}. Ambon Malay has five vowels, but these are best analyzed as bisyllabic than diphthongs \cite{68} and there are 19 consonants with very productive and common verbal reduplication. Being a Malay-based creole language like Kupang Malay, it has around 81\% of lexical similarity with Indonesian \cite{cahyawijaya-etal-2023-nusawrites}. 

\subsection{\textbf{Beaye}} 
Beaye (day\footnote{We used the \textbf{day} language code for Beaye because it lacks its own code, though it officially belongs to the Land Dayak language.}) is an understudied language spoken in Landak Regency in the northern region of West Kalimantan. Recent research has argued that the Beaye language belongs to the Benyadu-Bekati subgroup within the Land Dayak language family \cite{Sommerlot2020}. However, due to its status as a previously undescribed language, we don’t have any information about further classification of this language. Consequently, there are no exact estimations for the number of speakers and the use of language code in Beaye. The approximation of the speakers we obtained only from external source\footnote{https://budimiank.blogspot.com/2016/04/bahasa-angan-akan-masuk-kbbi.html}. We use Latin script because Beaye has no standard script. Beaye shares structural phonemic, and vocabulary similarities with Mali and Ba’aje, two other languages in the Benyadu-Bekati branch. Syntactically, Beaye adheres to Subject-Verb-Object word order, similar to Indonesian, but with a reduced verb affixation system. 

\subsection{\textbf{Kupang Malay}} 
Kupang Malay (mkn) is a Malay-based creole spoken in the western region of Timor Island and is used as a lingua franca (trade Malay) and for inter-ethnic communication. As a lingua franca Malay, indeed people around Kupang in Timor were not native speakers of Malay, however, through the creolization process the language evolved into a stable language making it different and separate from the family language \cite{Rafael2019}. Therefore, several dialects appeared, \textit{Air Mata, Alor Malay, and Basa Kupang} \cite{660784}, which should have been more than that because Kupang Malay speakers in fact come from a variety of different ethnic backgrounds absorbing and probably influenced by Vernacular languages associated with specific ethnic groups~\cite{Jacob2006,Lau2022}. Malay Kupang has limited morphologic, it has four prefixes but does not recognize suffixes, infixes, or confixes. Similar to Indonesian, Kupang Malay has a five-vowel system but without no schwa.  

\subsection{\textbf{Makassarese}}
Makassarese (mak) is a primarily local language besides Buginesse, spoken by about 2 million people in South Sulawesi. Makassarese has been used for centuries, since 1551 as the official language of the Kingdom (Gowa). It has three dialects which are Lakiung (Gowa), Turatea (Jeneponto), and Bantaeng (Maros-Pangkep). Makassarese morphology is characterized by highly productive affixation, such as prefixes, infixes, and suffixes, pure reduplication and partial reduplication, and widespread politicization of pronominal elements and specialty \cite{573169,44678}. One interesting thing about Makassarese is that the place of subject and predicate in Makassar usually changes (P-S pattern), similarly occurring in Western languages, especially Latin, or in Sanskrit and Arabic. Therefore, Makassarese lends a lot of vocabularies from Arabic \cite{Baso_2018}, and traditionally Makassarese includes two Indic-based scripts, a system based on Arabic, and various Romanised conventions. However, The Makassar script is no longer used \cite{cahyawijaya-etal-2023-nusawrites}.

\subsection{\textbf{Uab Meto}}
Uab Meto (\textit{uab}: language, \textit{meto}: land, upland, dry land), also known as Dawan, Meto, Timorese, Uab Pah Meto or Timor Dawan, is an Austronesian language spoken in West Timor and can be interpreted as a language who speak by people lived in “dry land” \cite{Liunokas_Adam_Lelo}. The majority of the native speakers are spread across Timor Island and the dialects variation as well, in particular South Central Timor, North Central Timor, Kupang, and Oecussi-Ambeno (called Baikeuno or Baikeno) regencies. Uab Meto consists of ten regional dialects \cite{Hajar2020,Liunokas_Adam_Lelo,Olbata2016/11}, which differ primarily in pronunciation and vocabulary. Uab Meto language comprises seven vowels, with one diphthong, and 11 consonants, including affixation and reduplication in its morphological process \cite{321495}. It has two key features: undergoes productive metathesis, and its verbs are always followed by subject markers, which vary based on the subject’s form, whether nominal or pronominal \cite{Budiarta2009}. 

\section{Data Creation}
Creating a low-resource language dataset is costly and time-consuming \cite{10.1162/tacl_a_00474}. To address this, we've adopted a volunteer-based approach inspired by participatory research methods \cite{Nekoto2020,winata-etal-2023-nusax}. This approach involves native speakers in our translation circle initiative. For data quality assurance, we conducted double-blinded evaluations by annotators and evaluators.

\subsection{Data Source}
In our study, we utilized two sources of data. The first source is the Tatoeba Dataset, a crowd-sourced collection of translations in multiple languages, including Indonesian to English pairs. We selected 4,000 sentences from this dataset to simplify the translation process. The second source of data is the NusaX Lexicon Dataset\footnote{https://github.com/IndoNLP/nusax}, which was used to create a lexicon for the Beaye language. This dataset comprises 2,442 rows, including 477 words in Indonesian-English pairs, meticulously annotated by professional annotators to ensure top-notch quality. 

\begin{figure*}
 \centering
 \includegraphics[scale=0.7]{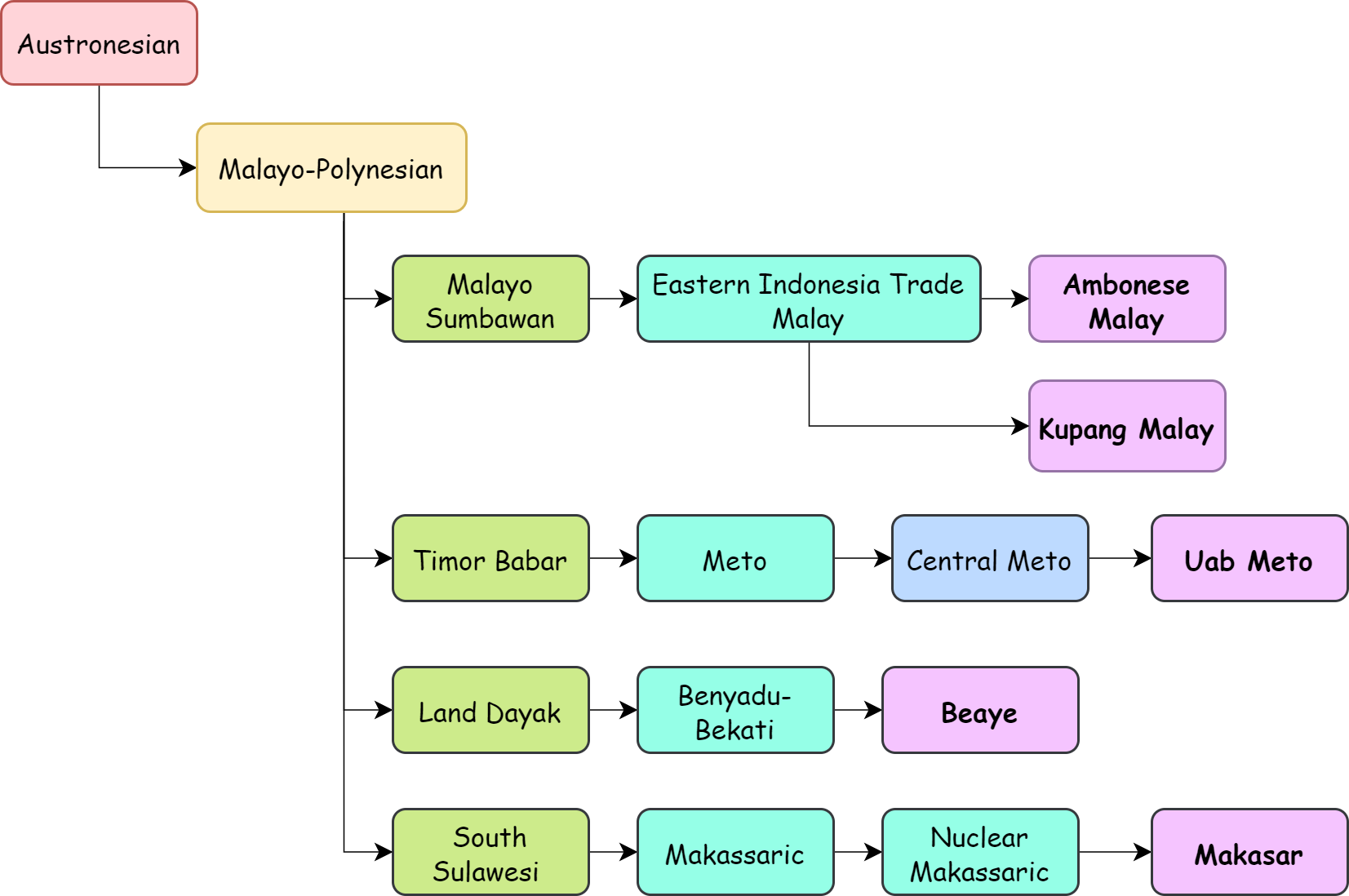}  
 \caption{Taxonomy of the languages}
\end{figure*}

\subsection{Translation-circle-initiative}
\textbf{Preliminary experiment} Before the initiation of our main study, we conducted a preliminary experiment to anticipate data collection challenges in our low-resource, volunteer-based initiative. We gathered data from readily accessible languages like Uab Meto and Kupang Malay, preparing around 500 sentences for native speakers of both languages to translate from Indonesian. The experiment had no time constraints, and we provided minimal instructions. As a result, we encountered difficulties as annotators found it challenging to translate all the data due to its size and time-consuming nature. Additionally, the media used did not fully meet the annotators' needs.

\textbf{Data Chunk} 
To address the challenges from our early experiment, we divide the English-Indonesian pair data into manageable chunks, aligning the quantity with the available annotators. Each chunk was assigned to annotators in respective languages for translation, fostering collaborative work and lightening the workload. Importantly, annotators and evaluators maintained anonymity, ensuring that they did not know each other.

\textbf{Evaluator Notes}
Our data is partitioned into segments, which are then randomly assigned to annotators. We maintain strict confidentiality, objectivity, and fairness by ensuring that annotators and evaluators are unaware of each other. Evaluators provide feedback and notes to us for consideration, covering aspects such as translation, study suggestions, and communication results during the assessment process.

\textbf{The Decision}
Ultimately, the final decisions are our responsibility, taking into account feedback from evaluators and communication with annotators. In the five less-studied languages, evaluators largely concurred with the annotators' findings. The main alterations involved dialect, spelling, and typographical errors, while other adjustments were minor and did not impact the overall dataset.

\subsection{Annotator Recruitment}
Annotating data in low-resource languages often relies on local language communities \cite{winata-etal-2023-nusax,Nekoto2020}, but finding such communities can be challenging. As a result, we collaborated with only one Ambonese community and had to enlist acquaintances and individuals introduced to us for data collection. Our specific criteria for annotators included a minimum 20-year residency in the understudied language areas, aiming to involve university students and young speakers engaged in language development. Furthermore, recruitment was done by considering several aspects.

\textbf{Annotator Compliance} was essential for the success of our study. Surprisingly, we received enthusiastic responses from native speakers across diverse backgrounds, including traditional elders, university students, graduate students, teachers, workers, and cultural custodians. However, some languages, like Beaye, posed challenges in finding volunteers due to their limited documentation. Nevertheless, we engaged 5 native speakers to assist with translation, evaluation, and lexicon compilation in Beaye.

\textbf{Annotator Quality} Our focus was on selecting high-quality annotators who met specific criteria, such as university students in their second year or beyond, university graduates, educators with language expertise, participants with long residence in the region, and cultural custodians\footnote{We employed cultural custodians solely for the Uab Meto language due to their vital role in preserving local linguistic nuances} with expertise in oral traditions. While some annotators lacked translation experience, we implemented a rigorous evaluation process, with most evaluators having translation expertise.

\textbf{Annotator Responsibility} We initially sought volunteers willing to take on translation responsibilities, but some had to discontinue due to unavoidable commitments. Out of approximately 40 individuals approached 36 and persevered till the end. To address this, we reached out to non-responders for recommendations to fill vacant positions in the translation task.

\begin{table*}
\centering
\begin{tabular}{lllll}
\hline
\textbf{lang} & \textbf{sent (avg)} & \textbf{vocab} & \textbf{readability} & \\
 & & & \textbf{mean} & \textbf{median} \\
\hline
Ambonesse Malay & 4.339 & 1773 & 2.745 & 2.9 \\
Kupang Malay & 4.501 & 1841 & 2.454 & 2.1 \\
Uab Meto & 4.714 & 2159 & -0.431 & -1.5 \\
Beaye & 4.357 & 2613 & 1.908 & 1.3 \\
Makassarese & 3.093 & 2095 & 6.284 & 5.6 \\
\hline
\end{tabular}
\caption{Statistics of languages understudy}
\end{table*}

\textbf{Lexicon Annotator} For the Beaye language, we pioneered the development of a bilingual lexicon by engaging two native speakers. One translated each word into all potential lexemes, while the other assessed the lexicon's quality. Evaluation outcomes were discussed thoroughly with both native speakers to determine appropriate translations, scrutinize for potential errors, and enhance lexicon quality. We also asked one linguist who is working on the beaye language to evaluate the annotated lexicon by the annotators.

\subsection{Translation Procedure \& Tools}
\textbf{Translation Procedure} Instructing annotators to ensure accurate and consistent translations, we emphasized retaining the original text's meaning, avoiding direct translation of entities like people, organizations, locations, and time. Our guidance focused on the languages' uniqueness. We also specified: \textbf{(1)} Using language-specific dialects; \textbf{(2)} Keeping the original sentence when an appropriate word is unavailable; \textbf{(3)} Preserving entities in the text; \textbf{(4)} Maintaining original punctuation and capitalization. Dealing with dialect variations, especially in Indonesian local languages, recruiting annotators, particularly in Uab Meto with its 10 dialects, presented challenges. As a result, we chose to work with just two dialects, Amanuban-Amanatun and Mollo-Miamafo, due to their accessibility and central role in the language.

\textbf{Translation Tools} Furthermore, In regions with infrastructure and accessibility limitations, we adapted our translation methods. We considered feedback from preliminary experiment respondents, especially concerning media translation. Our approach was tailored to match the technological capabilities and accessibility of annotators in different areas. This involved using various tools such as online resources, email, WhatsApp for document exchange in Excel or Word Format, and even physical printouts for annotators with limited technological access.

\section{Evaluation}
\subsection{Corpus Statistics}

Our parallel corpus consists of 18,000 sentences in total, with the details are 4 languages containing 4,000 sentences while the Makassarese only 2,000 sentences. Each translation has its Indonesian and English translation. The total number of annotators and evaluators who participated in the translation process is 36 people consisting of various groups and occupations. The average sentence length for all languages is approximately 4 words, however, there is also a language that has fewer words than other languages due to the size of the corpus. This shows that every sentence in Indonesian can be translated both literally or contextually in each pair of languages. For more details, Table 2 shows the statistics for each language. 

\subsection{Flesch-Kincaid Readability} We also measured the readability of each language and text simplification evaluation by using the Flesch-Kincaid readability method \cite{Farr1951}. Although, this method has not been used in the NLP field, especially as a text simplification evaluation metric, however, the components of Flesch-Kincaid and other related statistics can be used to help understand what corpus parallel are doing \cite{tanprasert-kauchak-2021-flesch}. Table 2 provided readability scores and linguistic characteristics which valuable insights into the accessibility and complexity of texts in various languages. It seems that Ambonesse Malay and Kupang Malay emerge as highly accessible languages than Uab Meto's Flesch-Kincaid, which has negative readability scores. The negative score implied that Uab Meto words are exceptionally simple and straightforward, potentially due to the basic vocabulary. Furthermore, Beaye presents another low-resource language with texts that are readily comprehensible and valuable for machine translation and localization. However, Makassarese tends to be a more challenging language in terms of readability, this complexity may pose difficulties for NLP tasks, warranting additional effort in developing language resources and tools tailored to Makassarese.

\subsection{Language Pairs Diversity}
We also measured and assessed the diversity and complexity of language using Shannon's Entropy-based method \cite{Shannon1951}. These measures provide insights into the distribution of words, the predictability of language patterns, and the lexical and syntactic complexity of pair languages \cite{Liu2022}. Figure 3 highlights the result of the variations in lexical diversity and unpredictability of word usage across different languages. Overall, some language pairs exhibit significant differences in vocabulary richness, while others show more similarity. Language pairs involving Makassarese consistently stand out with the highest entropy difference values. Makassarese appears to have a notably higher level of lexical diversity compared to other languages in the table. Meanwhile, several language pairs, such as Indonesia vs. Ambonesse Malay and Indonesia vs. Kupang Malay, show moderate differences in lexical diversity. This suggests that these languages have relatively similar vocabulary richness but still exhibit distinctions. Furthermore, some language pairs, such as Uab Meto vs. Beaye and Ambonesse Malay vs. Kupang Malay, have very low entropy difference values, indicating high similarity in word usage patterns and lexical diversity.

\begin{table*}[t]
\centering
\begin{tabular}{lllllll}
\hline
\textbf{Languages} & \textbf{Smooth 1} & \textbf{Smooth 2} & \textbf{Smooth 3} & \textbf{Smooth 4} & \textbf{Smooth 5} & \textbf{Smooth 7} \\
\hline
Ambonesse Malay & 0.121 & 0.311 & 0.204 & 0.108 & 0.147 & 0.178 \\
Kupang Malay & 0.126 & 0.318 & 0.209 & 0.115 & 0.155 & 0.187 \\
Uab Meto & 0.038 & 0.136 & 0.075 & 0.029 & 0.047 & 0.063 \\
Beaye & 0.268 & 0.489 & 0.384 & 0.259 & 0.313 & 0.351 \\
Makassarese & 0.022 & 0.082 & 0.044 & 0.015 & 0.026 & 0.035 \\
\hline
\end{tabular}
\caption{Average BLEU scores for the five languages using six different smoothing methods.}
\end{table*}

\subsection{Statistical MT}
We conducted an experiment using our parallel corpus for the translation task. Considering the size of the corpus data and low resource setting, we decided to use a simple translation model, IBM Model 1 \cite{10.5555/1734086}, which is a model based on a statistical approach. Firstly, we performed Word Alignment based on Relative Positions to initialize first-order transition probabilities. Secondly, we used Bi-gram Language Modelling with Laplace Smoothing and Backoff to generate coherent target text. Thus, to handle out-of-vocabulary (OOV) we integrated Laplace (add-one) smoothing to reduce the probability contributions for unseen word pairs. Finally, for model evaluation, we could not use metrics such as alignment precision, alignment recall, and alignment error because there is no gold-standard annotation for alignment in our parallel corpus, therefore we employed BLEU Metric \cite{papineni-etal-2002-bleu} to evaluate the result. The evaluation was measured using various smoothing and trained for 25 iterations per language. Table 3 shows average BLEU scores for each language with the source language being Indonesian. 

\textbf{Model Performance} The table shows that each language has different average BLEU scores, indicating variations in the translation quality for these languages. We found that the Beaye language had a significantly consistent average BLEU score across various smoothing methods compared to several other languages. This suggests that translation tasks for Beaye are closer to human reference translations compared to the other languages. This is also supported by the fact that Indonesia and Beaye have relatively lower lexical differences and vocabulary variety. Meanwhile, Uab Meto consistently showed lower BLEU scores across all smoothing methods, indicating translation challenges and the need for improvement in translation tasks for this language. One reason that made Uab Meto performs poorly is complex syntax and semantics, as well as linguistic diversity such as many dialect or variations. It is likely to be very difficult to do transfer learning, zero-shoot, or few-shoot using this language. 

Furthermore, Ambonesse and Kupang Malay appear to be similar in scores. Both languages, as well as Indonesian, share linguistic roots and are both based on Malay. It has many benefits in the NLP context such as allowing for effective transfer learning, enabling knowledge gained from one language to enhance performance in the other, and parallel corpora and language models can improve machine translation for both languages and beyond.

\begin{table*}
\centering
\begin{tabular}{llllll}
\hline
\textbf{ind} & \textbf{day} & \textbf{freq} & \textbf{equivalents} & \textbf{monolingual} & \textbf{crosslingual} \\
\hline
anggun & bait & 12 & kind, goodness, nice & 0.367 & 0.244 \\
pengecut & bot & 9 & horrible, horrifying, fear & 0.406 & 0.211 \\
mengerikan & jet & 8 & nightmarish, awful, ill & 0.390 & 0.166 \\
menahan & nakap & 6 & jailed, secured, catching & 0.341 & 0.108 \\
hilang & kubes & 6 & gone, killed, die & 0.441 & 0.290 \\
senang & awan & 5 & cosiness, pleased, great & 0.282 & 0.506 \\
tanya & sikan & 5 & inquire, ponder, query & 0.375 & 0.247 \\
menolak & aye & 4 & no, turn down, dismiss & 0.223 & 0.235 \\
teriakan & ngampak & 4 & squealing, shout, screaming & 0.459 & 0.377 \\
rampas & nangko & 4 & robber, hijacker, plunderer & 0.465 & 0.280 \\
\hline
\end{tabular}
\caption{Beaye lexicon informations in top-10 words}
\end{table*}

\textbf{Variability Across Language} We found translation challenges related to variability across languages. For example, in smoothing 2, there is a significant difference between languages such as Beaye has higher scores, meanwhile, Makassarese produced lower scores even across all smoothing methods. Makassar language and Uab Meto suggest the need for focused improvement in terms of the quality of available data. Considering the low resource status still be a problem, especially the issue of dialect variety which is a common problem in the Indonesian language, efforts to build specialized models and linguistic expertise may be required. 

\textbf{Smoothing Impacts} Although we only using Simple SMT, the performance of each language already shows good indications for further development. As with the use of smoothing, the choice of smoothing method has an important impact on translation quality. Further research and experimentation are recommended to identify the most effective smoothing method for each language. For example, smoothing 2 and smoothing 5 generally result in higher BLEU scores and may be preferred for certain languages.

\subsection{Lexical Phenomenon}
The Beaye bilingual lexicon contains 710 words paired with Indonesian and English. Approximately 21.12\% of these words cannot be directly translated into Indonesian, and the rest do not share common terms with Indonesian. Our detailed examination revealed a notable influence of Indonesian on Beaye vocabulary, exemplified by instances like \textit{yakin} (confident) translating to \textit{yaken} in Beaye. This influence underscores the need for further exploration of Beaye, considering its understudied status. Moreover, our analysis yielded significant insights within this lexicon.

\begin{figure*}
 \centering
 \includegraphics[scale=0.5]{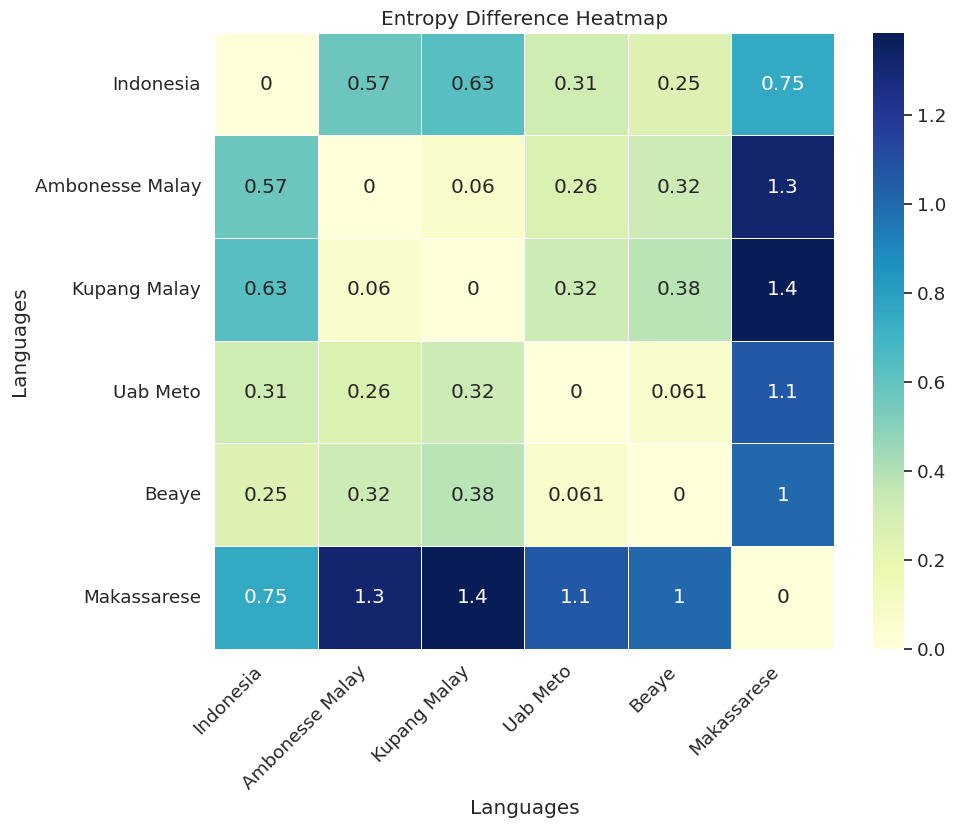}  
 \caption{Heatmap of variations in lexical diversity}
\end{figure*}

\textbf{Polysemy Phenomenon} We explored polysemy in the Beaye language, discovering that it often has multiple meanings for a single Indonesian word. For instance, the word \textit{bot} can be translated to \textit{cemas} (anxious), \textit{gentar} (afraid), or \textit{jelek} (awful). This polysemy poses challenges in NLP when dealing with ambiguous word representations, hindering accurate semantic understanding. This issue often arises due to word vector models representing all possible meanings as a single vector, leading to meaning conflation \cite{CamachoCollados2018}.

\textbf{Allophones Phenomenon} We're also exploring the Allophones Phenomenon in the Beaye language, where different sounds are used for the same word function \cite{silfverberg-hulden-2017-weakly}. An illustrative example is the term \textit{ayu'k} which translates to "big" in English. However, in comparative or superlative contexts like "enormous," "huge," "big," "bigger," "biggest," or "largest," it retains its use but with sound variations. In NLP, addressing allophones is closely tied to morphological segmentation, a crucial task in various NLP applications. Addressing this phenomenon can alleviate sparseness issues in NLP tasks, distinguishing negation morphemes from noun derivations and enhancing morphological analysis.

\textbf{Homonym Phenomenon} We also explored the Homonym Phenomenon, where words share the same phonetics but have distinct meanings and functions in both languages. For instance, the Indonesian word \textit{galak} can mean \textit{pemarah} (grumpy), \textit{buruk} (stern), \textit{ganas} (fierce), while in the Beaye language, it translates to \textit{keinginan} (desire). When developing an effective NLP model for the Beaye language, addressing homonyms is crucial to ensure accurate understanding and generation of text or speech.

\subsection{Lexical Similarity} We conducted an experiment to evaluate the semantic relationships between Indonesian and Beaye words. We utilized Fasttext word embeddings \cite{bojanowski2017enriching} to generate word representations from our parallel data. Subsequently, we measured word similarity using cosine similarity on the Indonesian-Beaye word pairs found in the Beaye lexicon.

\textbf{Monolingual Similarity} We calculated Fasttext similarity scores using only the Indonesian Fasttext model in our experiment. We didn't account for the nuances of the Beaye language but used it to assess similarity, which can be valuable for those with limited resources or a stronger focus on Indonesian. The results in Table 4 show that Indonesian words are generally quite similar to existing Beaye words. This simplifies implementation, reduces complexity, and suggests that the Beaye model may suffice for obtaining basic similarity scores in the language.

\textbf{Cross-Lingual Similarity} In the second experiment, we measured similarity using Fasttext models from both languages. The results were notably dissimilar, even more so than in monolingual scenarios. This highlights the substantial differences in the structures, vocabulary, or contexts of the two languages, making it challenging to align their embeddings effectively. As previously discussed, Beaye features both Polysemy and Homonymic phonemes, and the model's inability to distinguish these meanings can contribute to dissimilarities between word pairs. Additionally, both languages may exhibit variations in word usage, grammar, or syntax that are difficult to accurately capture in cross-language embeddings.

\subsection{Data Collection Challenges} 
Building a parallel corpus for languages with fewer and undocumented resources and speakers is complex and time-consuming. We encountered difficulties in finding participants who matched our criteria and standards. Furthermore, selecting excellent and accurate translations amidst the variety of dialects is another challenge we faced. Although we asked external annotators to consider using sentences that could be understood by speakers of different regional dialects, finding annotators with comprehensive knowledge of the various dialects in a language was difficult. Limited access to the internet and information technology is also a challenge in data collection and processing. This has led to the use of many types of media in translation, such as physical print, which slows down the data processing process and is difficult for annotators to access. We found it challenging to check the data and receive feedback from the annotators, so we often had to copy the translations into a digital format to facilitate checking and evaluation. 

\section{Conclusion and future work}
In this study, we have presented the essential steps and evaluation of Indonesian-Ambon Malay, Kupang Malay, Beaye, Makassarese, and Uab Meto parallel corpus. We also release first bilingual lexicon of Beaye, an under-documented language in Indonesia. The use of this parallel corpus and bilingual lexicon is expected to accelerate the process of NLP research in low resource settings. In addition, we have created an open repository on Github so that this parallel corpus can be freely and easily accessed by the general public and researchers. We hope that this parallel corpus can contribute positively to the preservation of local languages in Indonesia and become a reference for further research in related fields. In order to improve the accessibility of information on local languages in Indonesia, further steps will be taken to increase the number of regional languages. We also plan to test the parallel corpus using NLP advanced method dealing with low resource languages to build a multilingual translation model.
\paragraph{Acknowledgements:} We thank to Nur Islamiah, Abd. Rahman Sholeh, Ode Dermansya and 40 annotators who helped us in building the corpus. We also express our gratitude to those who have assisted in the review of this manuscript.

\


\bibliography{eamt24}
\bibliographystyle{eamt24}
\end{document}